# MAPPING SAVANNAH WOODY VEGETATION AT THE SPECIES LEVEL WITH MULTISPECRAL DRONE AND HYPERSPECTRAL EnMAP DATA

*Christina Karakizi[1,3], Akpona Okujeni[2], Eleni Sofikiti[3], Vasileios Tsironis[3], Athina Psalta[3], Konstantinos Karantzalos[3], Patrick Hostert[2], Elias Symeonakis[1]*

[1] Department of Natural Sciences, Manchester Metropolitan University, Manchester, M1 5GD, UK – c.karakizi@mmu.ac.uk
[2] Geography Department, Humboldt-Universität zu Berlin, Unter den Linden 6, 10099 Berlin, Germany
[3] Remote Sensing Laboratory, National Technical University of Athens, 15780 Zographos, Greece

## ABSTRACT

Savannahs are vital ecosystems whose sustainability is endangered by the spread of woody plants. This research targets the accurate mapping of fractional woody cover (FWC) at the species level in a South African savannah, using EnMAP hyperspectral data. Field annotations were combined with very high-resolution multispectral drone data to produce land cover maps that included three woody species. The high-resolution labelled maps were then used to generate FWC samples for each woody species class at the 30-m spatial resolution of EnMAP. Four machine learning regression algorithms were tested for FWC mapping on dry season EnMAP imagery. The contribution of multitemporal information was also assessed by incorporating as additional regression features, spectro-temporal metrics from Sentinel-2 data of both the dry and wet seasons. The results demonstrated the suitability of our approach for accurately mapping FWC at the species level. The highest accuracy rates achieved from the combined EnMAP and Sentinel-2 experiments highlighted their synergistic potential for species-level vegetation mapping.

*Index Terms*— Land Degradation, Bush Encroachment, Machine/Deep Learning, Sentinel-2, Regression

## 1. INTRODUCTION

The savannah biome covers 50% of the African continent and 20% of the global land surface, while also representing around a third of terrestrial net primary production, and comprising a critical regulating component of the land carbon sink [1]. Savannahs are under threat from land degradation, not least due to woody vegetation encroachment [2], [3]. Monitoring the woody component of savannah vegetation is essential for identifying areas where degradation mitigation measures are a priority in order to achieve the UN Sustainable Development Goal (SDG) for Land Degradation Neutrality (LDN) by 2030 [4], [5], [6]. Monitoring the species composition of the woody component is also essential for land degradation assessments, while respective spatially distributed species composition maps are still unavailable.

Despite the recent breakthroughs in data availability, machine/deep learning (ML/DL) techniques and distributed cloud computing platforms, the ability to discriminate between woody species has been limited mainly to the field scale. For larger areas but still at regional scales, savannah woody species classification has been achieved employing very high spatial resolution imagery, as in spaceborne multispectral data [7], airborne hyperspectral (HS) combined with LiDAR data [8], as well as the integration of all three data types [9]. However, the high cost of such approaches has so far impeded their use for continuous monitoring and mapping over extended areas.

The recent launch of the Environmental Mapping and Analysis Program (EnMAP) that provides HS data with global coverage and a 30m-pixel resolution, could significantly improve our ability to distinguish between different vegetation species at larger scales. EnMAP is a newly launched German HS satellite mission that monitors and characterizes Earth's environment on a global scale, providing around 90 bands in VNIR and around 130 in SWIR parts of the spectrum [10], [11]. Previous studies have shown the value of EnMAP-like spaceborne imaging spectroscopy for generating vegetation type/species maps [12], [13] and capturing the variation in forest biomass related to species composition [14]. However, such approaches have never been tested in savannahs.

To this end, this work targets the accurate mapping of fractional cover of different woody vegetation species in a savannah region of South Africa (SA) based on EnMAP HS imagery from the dry season. Very high-resolution multispectral drone data are employed to produce training samples for fractional woody cover (FWC) regression experiments with the EnMAP imagery. The performance of different ML regression algorithms is benchmarked, while the contribution of additional multitemporal information derived from Sentinel-2 (S2) spectro-temporal metrics of both the dry and wet seasons is also assessed.

## 2. MATERIALS AND METHODS

### 2.1. Study Area and Woody Species

The study area is located in the western part of the Kagisano-Molopo municipality in the Northwest Province of South Africa, near the border with Botswana. It forms part of the South African Kalahari Desert and covers approximately 10,000 km². This region is characterized as a hot, semi-arid steppe, receiving the majority of its rainfall during the warmer summer months from October to March. Average midday temperatures range from 18°C in June to 31°C in January. The area typically receives about 400 to 600mm of rain per year, with an average annual precipitation around 450mm. [15]. The entire study area falls within the Savannah Biome dominated by the *Molopo Bushveld* vegetation type [16]. The main woody species in the study area are *Grewia flava, Senegalia mellifera, Vachellia erioloba, Vachellia luederitzii* and *Boscia albitrunca.* Amongst these, the *Senegalia mellifera* thorn tree is considered by locals to be one of the most encroaching and undesirable species. It can rapidly form impenetrable thickets and is not palatable to most animals.

### 2.2. Field Campaigns and Training Data Creation

Field data on vegetation types, with a focus on woody species, were acquired during in-situ campaigns over three sites in our study area during the dry season in late June 2023. Specifically, field campaigns included the annotation of savannah vegetation types using GPS measurements and geo-tagged images. Additionally, drone data were collected at 2cm spatial resolution using a DJI Mavic 3M multispectral camera, capturing RGB, Red-Edge, and Near-Infrared bands. Farmers and local experts participated in the field campaigns, providing our team with significant information regarding the encroaching species and the history of bush encroachment management and practices applied to the studied sites.

To create training data, the field annotations were combined with the drone data using a U-NET semantic segmentation architecture in order to produce labelled maps. The woody species annotated classes were grouped into three categories based on agronomical and spectral criteria: the *Grewia flava* class, the *Senegalia mellifera* class, and plants from the *Vachellia* genus family. The produced maps also included two non-woody species categories, Grasses and Soil, as well as a Shadow class. The very high-resolution classified maps from the drone mosaic were then used to generate FWC samples per woody species class at EnMAP's spatial resolution of 30m.

### 2.3. EnMAP and Sentinel-2 Data

Atmospherically corrected (L2A) EnMap data from two orbit stripes, were ordered via the EnMAP Instrument Planning Portal. The employed EnMAP imagery was acquired on August 2023, i.e., during the summer dry season in the area, when the field campaign data were also collected. All EnMap bands, except from those flagged as 'bad' from the metadata files, were kept for further processing.

In order to also assess the contribution of multitemporal information on an intermediate, between the two main datasets, spatial resolution, S2 10-m data of multiple dates, were also employed. The Framework for Operational Radiometric Correction for Environmental monitoring (FORCE; [17]) was utilized for downloading and pre-processing the S2 data. Spectro-temporal metrics ($10^{th}$, $25^{th}$, $50^{th}$, $75^{th}$, $90^{th}$ percentiles) from the ten 10/20-m S2 bands and three spectral indices, NDVI, EVI and SAVI, were calculated for dry and wet season of 2023.

### 2.4. Fractional Woody Cover Mapping at Species Level

FWC regression experiments at species level were carried out using the EnMAP-Box, both as a QGIS plugin and directly from python commands. The EnMAP-Box offers advanced functionalities to visualize and process hyper- and multispectral, and multitemporal remote sensing data, it implements novel spectral libraries concept, and provides easy access to published algorithms from different fields of environmental research [18]. Four commonly used ML algorithms [19], [20], [21], i.e., Random Forest (RF), Support Vector Machine (SVM) with a Radial Basis Function, Kernel Ridge (KR) and XGBoost (XGB) were employed along with the EnMAP data via the EnMAP-Box functionalities. The contribution of multitemporal information from the S2 data was also assessed by including the calculated spectro-temporal metrics as additional predictors in the regression experiments.

## 3. RESULTS & ANALYSIS

The neural network classification of the drone imagery using the U-NET architecture achieved very high accuracy levels, exceeding 89% for all three sites. In Figure 1, the top image shows the drone aerial mosaic for Site #1 in a natural colour composite, while the classified map is displayed at the bottom. At this site, a manual stem-burn procedure was used to manage *Senegalia mellifera* encroachment, whereas the field south of the road remains unmanaged. The classified map depicts the *Senegalia mellifera* land cover class in yellow, clearly showing a significant reduction of bushes in the managed northern part compared to the unmanaged southern part.

The classified drone maps from all three sites were used to produce FWC samples at the species level, which served as input for the regression experiments. Four different regressors were benchmarked for FWC estimation using EnMAP HS dry-season single-date data, followed by experiments combining these data with multitemporal spectro-temporal S2 metrics from both dry and wet seasons.

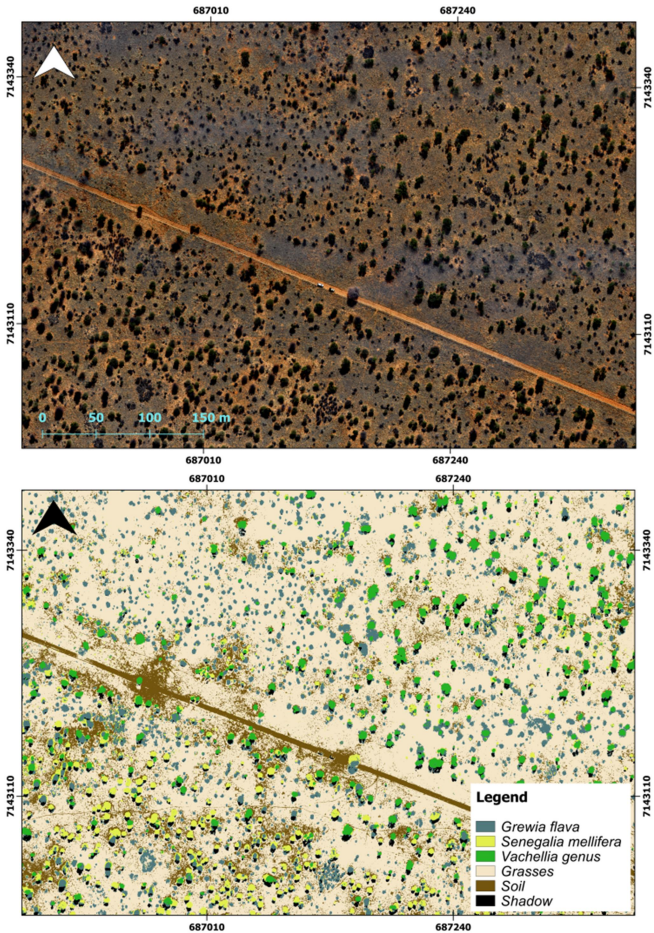

**Figure 1**: The drone aerial mosaic for Site #1 in a natural color composite (top) and the respective classified map produced by the U-NET DL architecture (bottom).

**Table 1.** Per Species RMSE rates for the FWC regression experiments at 30m, using EnMAP (EnM) data and their combination with S2 spectro-temporal metrics (EnM + S2). Lowest RMSE rates per column are marked in bold.

| Regr. Alg. | *RMSE (%)* | | | | | |
|---|---|---|---|---|---|---|
| | *Grewia flava* | | *Senegalia mellifera* | | *Vachellia* genus | |
| | EnM | EnM + S2 | EnM | EnM + S2 | EnM | EnM + S2 |
| **RF** | **3.86** | 3.29 | 4.60 | 3.77 | 3.96 | **3.36** |
| **SVM** | 4.47 | 3.11 | 4.20 | 3.36 | 4.07 | 3.80 |
| **KR** | 4.27 | **3.07** | **4.12** | **3.35** | **3.91** | 3.58 |
| **XGB** | 3.99 | 3.36 | 4.76 | 3.98 | 4.30 | 3.68 |

Root Mean Squared Error (RMSE) rates for the performed experiments are presented in Table 1. In general in all cases, it is considered that RMSE rates remain quite low. Additionally, as observed, among the different regression algorithms, best performance for most experiments was recorded for the KR algorithm. The positive influence of the inclusion of multitemporal S2 information is apparent on the error rates of all combinations of algorithms and studied species.

In our study area, various stakeholders consider *Senegalia mellifera* to be the most undesirable encroaching species, so further analysis hereon is focused on this tree species. In Figure 2, the KR regression scatterplot for the combined EnMAP and S2 experiment for FWC mapping of *Senegalia mellifera* is presented. Larger prediction errors are spotted for larger FWC rates, while in most cases FWC does not exceed 30% of the 30-m pixels. For this experiment, the coefficient of determination values (i.e., $R^2$) reached a high rate of 76.25%. For the same data combination, the other regressors performed as follows: RF with an $R^2$ of 69.88%, SVR with an $R^2$ of 76.00% and XGB with an $R^2$ of 66.46%.

Figure 3 presents subsets of the *Senegalia mellifera* FWC map produced for our study area from the KR experiment employing both EnMAP and S2 data. Similarly to the scatterplot of the same experiment, the map also shows that FWC at the EnMAP spatial resolution, i.e., 30m, does not exceed the 30% rate, which is representative for woody species in savannah environments. High FWC rates of *Senegalia mellifera* (yellow color) are observed over regions of unmanaged fields and rangelands, while lower levels (blue color) are spotted over managed fields, roads and salt pans. In the lower map of Figure 3, a checkerboard pattern of FWC can be observed close to Sites #2 and #3. For this particular field, woody vegetation encroachment management has been applied through aerial spraying in perpendicular flight paths, creating cleared regions and leaving square encroached patches intact.

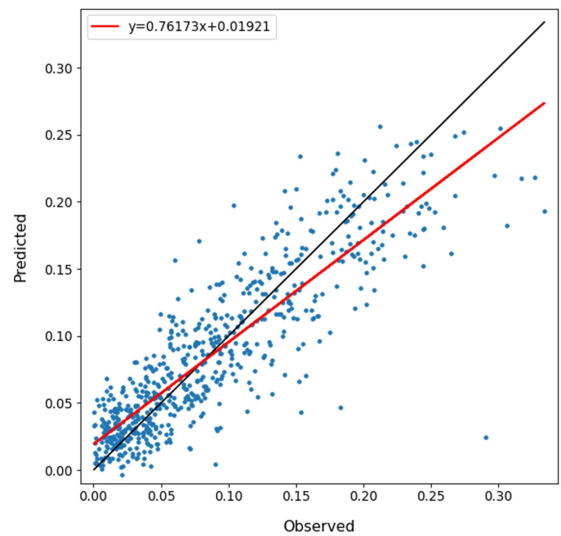

**Figure 2**: The KR regression scatterplot of the combined EnMAP and S2 experiment for mapping the FWC of *Senegalia mellifera*. $R^2$ reached a high 76.25% rate.

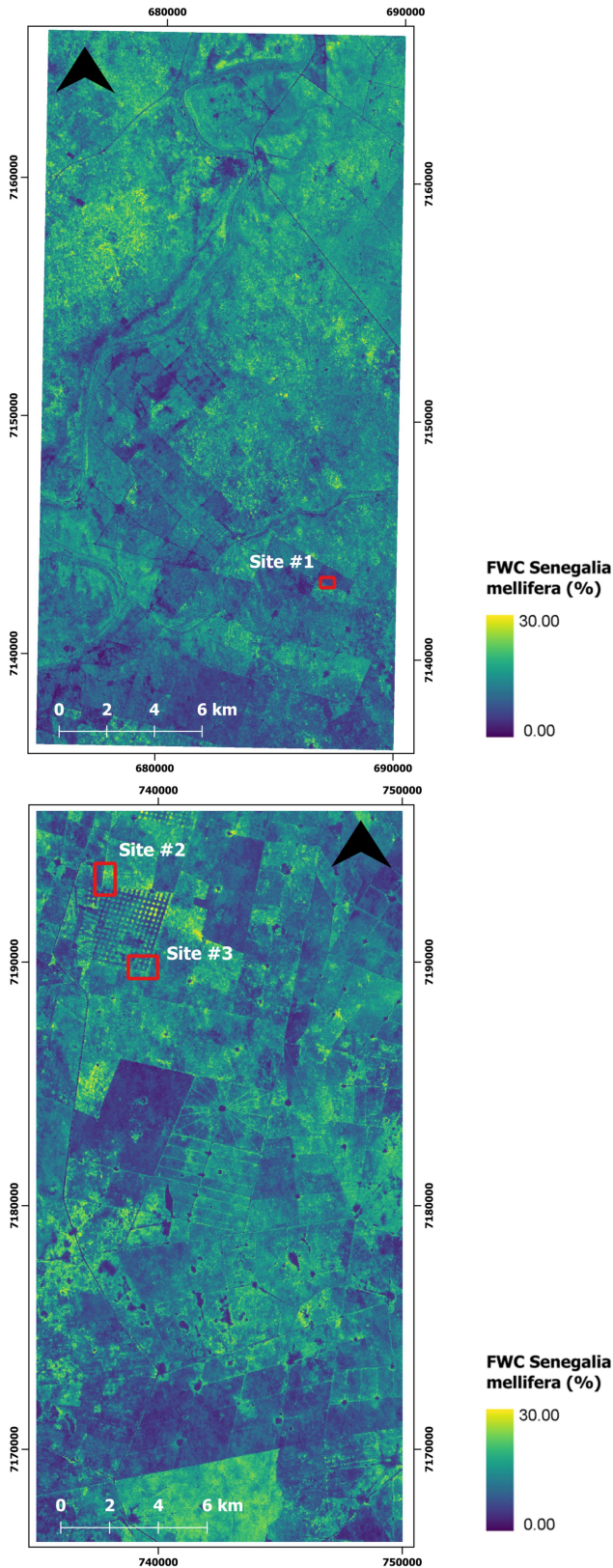

**Figure 3:** Subsets of the *Senegalia mellifera* FWC map from the KR experiment employing both EnMAP and S2 data. Field campaign sites are demonstrated with red boxes.

## 4. DISCUSSION & CONCLUSIONS

Savannahs in Africa are vital ecosystems that support millions of pastoralists whose livelihoods are increasingly threatened by woody encroachment. This encroachment involves the proliferation of bushes that displace palatable grasses, leading to significant economic hardship. Pastoralists not only lose livestock due to the reduced availability of grazing land but also incur substantial costs in managing the encroached land and attempting to halt the progression of this encroachment. Earth Observation efforts to monitor and map savannah encroachment have primarily focused on producing land cover or fractional cover maps that distinguish between woody and grass components [2], [3], [22], [23], [24], [25]. Communication with local stakeholders and farmers has revealed a crucial aspect of woody encroachment: not all bushes have a negative impact on the land's services. Therefore, monitoring woody cover at the species level is essential for realistic assessments of savannah land degradation and holds significant importance for local populations.

Towards this direction, the developed pipeline evaluates the capabilities of the new-generation imaging spectrometer EnMAP for monitoring and mapping savannah woody vegetation at the species level. The proposed methodology is designed to produce fractional woody cover (FWC) maps for three representative woody species families in a savannah region of South Africa (SA) at a 30-m spatial resolution, based on EnMAP dry season imagery. Multispectral drone data have been proven well-suited for upscaling species-level field data to spaceborne scales to produce FWC samples at 30m. All benchmarked regression algorithms presented relatively low prediction errors, with the Kernel Ridge (KR) regressor achieving the highest accuracy rates for most experiments. Adding Sentinel-2 spectro-temporal metrics from both dry and wet seasons as extra predictors in combined regression experiments significantly enhanced FWC mapping accuracy.

A thorough visual inspection of the FWC maps and low error rates from the accuracy assessment demonstrated the suitability of our approach for accurately mapping FWC at the species level. Future work could explore the full synergistic potential between the two sensors by employing EnMAP data from multiple dates in both seasons. Overall, the evaluation and derived conclusions raise significant expectations for expanding such methodologies towards operational monitoring and mapping over extended areas, providing essential information to intergovernmental and governmental policymakers and local stakeholders.

## 5. ACKNOWLEDGEMENTS

This work has received funding from the European Union's Horizon 2020 research and innovation programme under the 2019 Marie Skłodowska-Curie Individual Fellowships grant agreement SAV-EO No 894403.